\documentclass[aps,prl,twocolumn,showpacs,floatfix,superscriptaddress]{revtex4-1}

\usepackage{bm}
\usepackage[normalem]{ulem}
\usepackage{soul}
\setstcolor{blue}
\usepackage{times,mathrsfs}
\usepackage{graphicx,color}
\usepackage{amsmath,amssymb}
\usepackage[colorlinks,citecolor=red]{hyperref}
\usepackage[capitalize]{cleveref}

\begin{document}

\title{Resolution and Relevance Trade-offs in Deep Learning}

\author{Juyong Song}
\affiliation{Asia Pacific Center for Theoretical Physics, Pohang, Gyeongbuk 37673, Korea}
\affiliation{Department of Physics, Pohang University of Science and Technology, Pohang, Gyeongbuk 37673, Korea}
\affiliation{The Abdus Salam International Centre for Theoretical Physics, Strada Costiera 11, 34014 Trieste, Italy}

\author{Matteo Marsili} \email{marsili@ictp.it}
\affiliation{The Abdus Salam International Centre for Theoretical Physics, Strada Costiera 11, 34014 Trieste, Italy}

\author{Junghyo Jo} \email{jojunghyo@kias.re.kr}
\affiliation{Asia Pacific Center for Theoretical Physics, Pohang, Gyeongbuk 37673, Korea}
\affiliation{Department of Physics, Pohang University of Science and Technology, Pohang, Gyeongbuk 37673, Korea}
\affiliation{School of Computational Sciences, Korea Institute for Advanced Study, Seoul 02455, Korea}

\date{\today}

\begin{abstract}
Deep learning has been successfully applied to various tasks, but its underlying mechanism remains unclear.
Neural networks associate similar inputs in the visible layer to the same state of hidden variables in deep layers.
The fraction of inputs that are associated to the same state is a natural measure of similarity and is simply related to the cost in bits required to represent these inputs. 
The degeneracy of states with the same information cost provides instead a natural measure of noise and is simply related the entropy of the frequency of states, that we call {\em relevance}. Representations with minimal noise,  at a given level of similarity (resolution), are those that maximise the relevance. 
A signature of such efficient representations is that frequency distributions follow power laws. 
We show, in extensive numerical experiments, that deep neural networks extract a hierarchy of efficient representations from 
data, because they {\em i)} achieve low levels of noise (i.e. high relevance) and {\em ii)} exhibit power law distributions. 
We also find that the layer that is most efficient to reliably generate patterns of training data is the one for which 
relevance and resolution are traded at the same price, which implies that frequency distribution follows Zipf's law.
\end{abstract}

\maketitle

\section{Introduction}
Deep learning (DL) is similar to physics modeling in the sense that it extracts relevant features from data
that can be used for discriminating and generating new data~\cite{Byers2017}.
DL has state-of-the-art performance in various fields, including image/voice recognition and language translation~\cite{Lecun2015}, and it has 
also been applied to learn phases and phase transitions in physics~\cite{Carrasquilla2017}.
DL works by extracting a hierarchy of features, in which primitive features are extracted in shallow layers near the input layer,
while more abstract features emerge in deeper layers after processing the primitive features~\cite{bengio2013representation, Lecun2015}.
However, why DL works so well remains unclear~\cite{mehta2014exact, tegmark2017}. Besides the observation that the extracted features
are clearly relevant in most cases, no principled explanation for its success has emerged.

\begin{figure}[t]
\begin{center}
\includegraphics[width=8.7cm]{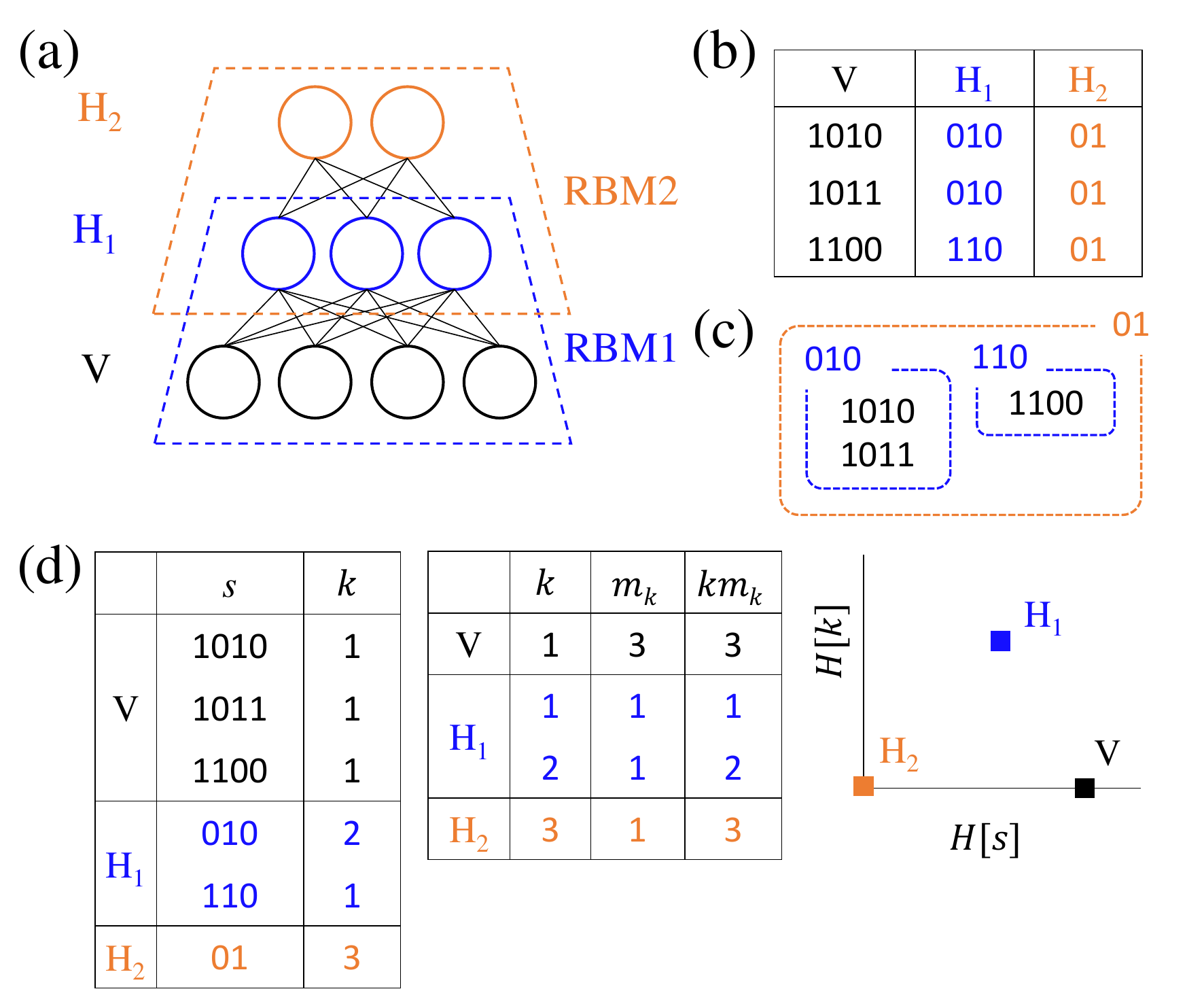}
\end{center}
\caption{ \label{Figure1}
Information processing in deep learning.
(a) A Deep Belief Network (DBN), consisting of one visible layer ($V$) and two hidden layers ($H_1$ and $H_2$),
is composed of stacks of restricted Boltzmann machines (RBMs).
(b) An example of data representation in the DBN.
The DBN maps three input data in $V$ to hidden states in $H_1$ and $H_2$.
(c) The data representation can be considered to be a hierarchical data grouping based on the hidden states.
The forward propagation of input data to deep layers is a coarse-graining process.
Subsets of distinct states on the shallow layers are transformed to identical states in deep layers.
(d) Then, $V$, $H_1$, and $H_2$ have different sets of distinct states $s$.
Two entropies are obtained for each layer on the basis of
the frequency $k$ of distinct states and its degeneracy $m_k$:
$H[s]$ represents the uncertainty of state distinguishability,
and $H[k]$ represents the uncertainty of state frequency.
}
\end{figure}

In this study, we try and contribute to the understanding of DL, by seeing it as 
an effective data grouping method~\cite{bengio2009learning, bengio2013representation, chen2015deep},
in which information is propagated from the input layer through progressively deeper layers. 
Each layer of the architecture extracts a compressed representation of the inputs, at coarser and coarser resolution as one moves deeper in the network. 
This compressed representation is achieved by mapping each input $\bm{v}$ in the visible layer, to one of the states $\bm{s}_{\bm{v}}$ of the hidden variables of that layer  (see Fig. \ref{Figure1}) \footnote{The distribution $P(\bm{h}|\bm{v})$ of states in the hidden layer, for a fixed input, was found to be sharply peaked in all examples we studied, which makes it reasonable to approximate it with a singleton, as in hard clustering.}. The inputs that correspond to the same state $\bm{s}$ supposedly share similar features. A key observation is that {\em the only quantitative measure of similarity available to the network is 
the  number $k_{\bm{s}}$ of inputs that correspond to a state $\bm{s}$}: An input that is represented by a state $\bm{s}$ with a large  $k_{\bm{s}}$ is expected to have very generic features, which are well characterised at that layer, whereas inputs with more specific features that are not well resolved, will fall into states with low $k_{\bm{s}}$. This similarity is measured by the information cost $E({\bm{s}})=-\log (k_{\bm{s}}/M)$, which is the number of bits required to store each input which corresponds to state $\bm{s}$. Following a recently suggested analogy with statistical mechanics~\cite{mora2011}, we'll also call $E({\bm{s}})$ energy bearing in mind that its precise meaning is that of an information cost. 
A natural measure of resolution of the representation achieved at a given layer is given by the average information cost
\begin{equation}
\label{Hs}
H[{s}]=\langle E \rangle=-\sum_{\bm{s}}\frac{k_{\bm{s}}}{M}\log\frac{k_{\bm{s}}}{M},
\end{equation}
where $M=\sum_{\bm{s}} k_{\bm{s}}$ is the number of inputs. 
Assuming that all inputs are different, the visible layer corresponds to a state of maximal resolution $\langle E \rangle=\log M$, whereas a layer, where all inputs are mapped to the same state, would correspond to minimal resolution $\langle E \rangle=0$. 
A layer with an intermediate resolution $0< \langle E\rangle< \log M$ will feature a distribution of ``energy levels'', with states with ``low energy'' states that are described in terms of well characterized features, and poorly characterized states with ``high energy''.

Further, we argue that there are distinctive statistical properties that make the representation at each layer of the hierarchy optimal. While low energy (i.e. well populated) states, correspond to well characterised inputs that are well discriminated from the others, 
poorly sampled states correspond to poorly characterised inputs, that are not yet discriminated from each other, at resolution $\langle E\rangle$. For example, inputs that are classified in states with $k_{\bm{s}}=1$ are not necessarily different from the others, rather they likely correspond to inputs that stand out because they are particularly noisy. 
We remark that {\em the degeneracy of energy levels provides the only quantitative measure of the uncertainty at a particular resolution $E$}. The level of noise of a particular state $\bm{s}$ can only be quantified in terms of the number of states with the same energy $E=E({\bm{s}})$. It is then natural to take, as a measure of noise of the representation, the (log) of the number of states with a given energy $E$, i.e. the entropy $S(E)$. 
The accuracy of the representation of a layer at a given resolution $\langle E\rangle$ can then be quantified by the average entropy $\langle S\rangle$, which is simply related to the {\em relevance}  \cite{marsili2013,marsili2015}
\begin{equation}
\label{SE}
 H[k]= -\sum_k \frac{km_k}{M}\log \frac{k m_k}{M}=\log M-\langle S\rangle,
\end{equation}
where $m_k$ is the number of states with energy $E=-\log (k/M)$. 
We conjecture that {\em optimal representations are those for which $\langle S\rangle$ is minimal, at a fixed resolution $\langle E\rangle$}. Put differently, optimal representations are those that maximise the relevance $H[k]$ at each level of resolution $H[s]$.

Within this picture, a feed forward network with many layers extracts a hierarchy of optimal representations with decreasing $\langle E \rangle$ (i.e. decreasing resolution $H[s]$) as one moves deeper and deeper in the architecture. The level of resolution $\langle E \rangle$ at each layer is constrained by the number of hidden variables, in principle. In practice, it is decided in the learning process in an unsupervised manner, and it ultimately depends on the data. Structureless (e.g. random) data is not expected to display features at many resolution scales, whereas data with a non-trivial structure may exhibit a rich hierarchy of features, spanning the resolution scale in a dense manner.

The rest of this paper focuses on exploring the consequences of the  conjecture above and in presenting evidence on numerical experiments.
We shall first review the definition of Deep Belief Networks (DBN), 
Then, we take the MNIST dataset~\cite{lecun1998mnist} as a benchmark, and show that, when a DBN is trained on the data, it extracts a sequence of representation at different layers that span uniformly the resolution scale. On the contrary, the representation of the data before learning (i.e. with random weights) or of structureless data (i.e. random or reshuffled data) concentrates on the upper or lower ends of the resolution scale. As shown in previous studies~\cite{marsili2013,marsili2015}, a clear signature of representations that maximise $H[k]$ at a given resolution $H[s]$, is that $m_k\sim k^{-\beta-1}$ follows a power-law behaviour. We find that learned representations indeed exhibit a power-law behaviour in $m_k$, with a gradually decreasing exponent as one moves deeper and deeper in the network. Different clustering methods (e.g. k-means) produce representations which do not feature power law cluster size distributions, indicating that power law distributions are not a characteristic of the data but rather of the mechanism of DL. Un-structured data or non-optimized networks instead are not characterised by power-law distributions of frequencies in general (although we observe power-law distributions in some cases). They also lack a rich hierarchical structure across the resolution scale.

Finally, we address the issue of optimal representation of the inputs. 
DL aims at striking a balance between compression and accuracy. Representations at very low resolution fall short of the necessary details for reconstructing the whole range of inputs, whereas representations at very high resolution include too many of these details. 
Our picture has the virtue of mapping both resolution and accuracy -- that is quantified by $H[k]$ -- on an information scale.
It therefore allows us to locate the point where a bit in resolution is traded for exactly one bit in accuracy. This turns out to be the point where {\em i)} the network displays the best generalisation ability and {\em ii)} the distribution of frequencies follows Zipf's law~\cite{newman2005}: $m_k\sim k^{-2}$. We conclude with a discussion of the results and of their implications.

\section{Results}
Among the various DL models, we adopted the DBN, a representative energy-based generative model~\cite{hinton2006reducing, hinton2006fast}.
A DBN is composed of stacks of restricted Boltzmann machines (RBMs) (Fig.~\ref{Figure1}a).
Each RBM consists of one visible layer and one hidden layer with restricted connections,
i.e. visible nodes are not connected to other visible nodes and hidden nodes are not connected to other hidden nodes.
Thus, nodes in the same layer are indirectly connected through the nodes in the neighboring layer within an RBM stack.
Given a visible and hidden state ($\bm{v}, \bm{h}$), the RBM defines an energy function,
\begin{equation}
E(\bm{v},\bm{h};\theta) \equiv - \bm{v}^{\top} W \bm{h}  - \bm{v} \cdot {\bm{a}} - \bm{h} \cdot {\bm{b}},
\end{equation}
where $\theta \equiv (W, \bm{a}, \bm{b})$ is the model parameters.
Specifically, the matrix $W$ represents the symmetric coupling strengths between the visible and hidden nodes,
and the vectors $\bm{a}$ and $\bm{b}$ control the biases of the visible and hidden states.
Then, a certain state ($\bm{v}, \bm{h}$) has the following probability,
\begin{equation}
P(\bm{v},\bm{h} ;\theta) = \frac{\exp(-E(\bm{v}, \bm{h} ;\theta))}{Z(\theta)},
\end{equation}
with $Z(\theta) = \sum_{\bm{v}', \bm{h}'} \exp(-E(\bm{v}',\bm{h}';\theta))$.
Hereafter, for brevity, we omit $\theta$ from the equations unless necessary.
Disconnection between the nodes in the same layer allows the RBM to factorize the probability
\begin{equation}
P(\bm{v}, \bm{h}) = \prod_{i} P(h_i|\bm{v}) P(\bm{v}) = \prod_{j} P(v_j|\bm{h}) P(\bm{h})
\end{equation}
to conditional probabilities.
We use the first equation to generate hidden states for given visible states
and the second equation to generate visible states for given hidden states.
The forward and backward propagations are stochastic,
and repeated propagation achieves Gibbs sampling for the hidden and visible layers, respectively~\cite{huang2017accelerated}.

We first need to determine the parameter $\theta$ that can reliably reproduce data $\bm{v}$.
Suppose we have $M$ data points, $\{\bm{v}^\mu\}_{\mu = 1}^{M}$,
and each dataset has $N$ components, $\bm{v}^{\mu} = (v_1^{\mu}, \cdots, v_N^{\mu})$.
For independent datasets, the data likelihood given $\theta$ is ${\cal L}(\theta)=\prod_\mu P(\bm{v}^\mu; \theta)$.
Then, the RBM gives the following log-likelihood,
\begin{equation}
\log {\cal L}(\theta) = \sum_{\mu=1}^M \log \sum_{\bm{h}} P(\bm{v}^\mu, \bm{h}; \theta),
\end{equation}
through marginalization for all possible hidden states $\bm{h}$.
Learning through the Boltzmann machine algorithm optimizes $\theta$ by maximizing the log-likelihood ~\cite{ackley1985}.
After the learning is completed, we propagate the input data $\{\bm{v}^{\mu}\}_{\mu=1}^M$ forward to the first hidden layer
and obtain $M$ hidden states $\{\bm{h}^{\mu}_1\}_{\mu=1}^M$.
Here, we denote the visible and the first hidden layer as $V$ and $H_1$, respectively.
The hidden states $\{\bm{h}^{\mu}_1\}_{\mu=1}^M$ for $H_1$ serve as the input data for the second hidden layer $H_2$ (Fig.\ref{Figure1}a).
Then, we optimize $\theta$ for the second RBM stack and repeat the training for the remaining RBM stacks {(see Supplementary Text 1 for the details of the simulations)}.

After learning is completed, the DBN transforms $\{\bm{v}^{\mu}\}_{\mu=1}^M$ to $\{\bm{h}^{\mu}_1\}_{\mu=1}^M$,
$\cdots$, $\{\bm{h}^{\mu}_{\ell}\}_{\mu=1}^M$, $\cdots$ in the hidden layers. Distinct states on the shallow layers are transformed to identical states in the deep layers (Fig.\ref{Figure1}). 
Then, we count the frequency of state $\bm{s}$
\begin{equation}
k(\bm{s})=\sum_{\mu=1}^M \delta_{\bm{s}, \bm{h}^{\mu}}
\end{equation}
and the number $m_k$ of states with $k_{\bm{s}}=k$, in terms of which we compute $H[s]=\langle E\rangle$ and 
$H[k]=\log M-\langle S\rangle$ from Eqs. (\ref{Hs},\ref{SE}).

\begin{figure*}[t]
\begin{center}
\includegraphics[width=18.cm]{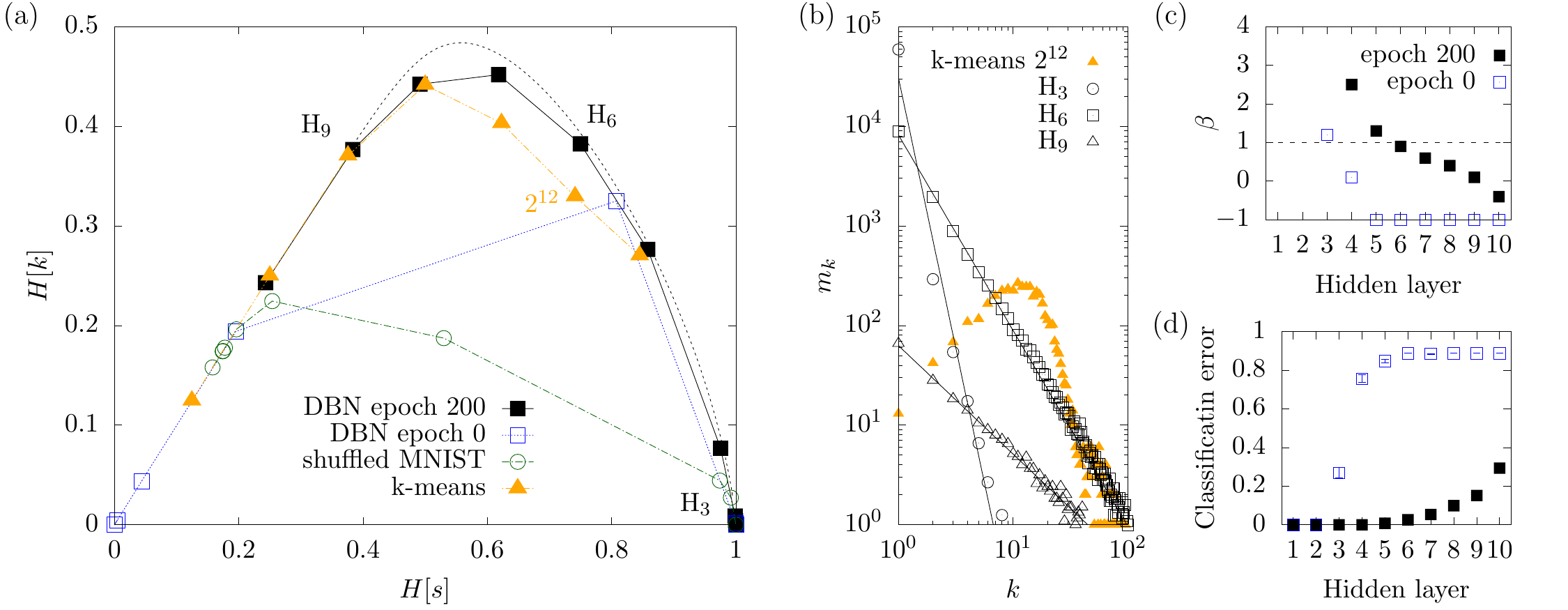}
\end{center}
\caption{\label{Figure2}
Critical data grouping of deep learning.
(a) The state entropy $H[s]$ and frequency entropy $H[k]$ of the hidden states for different layers
after a Deep Belief Network (DBN) learns the MNIST hand-written digit data (DBN epoch 200, filled black squares),
and before the DBN optimizes its parameters (DBN epoch 0, empty blue squares).
As a control, the two entropies are obtained for a structureless data (shuffled MNIST, empty green circles) in which pixels of MNIST digit images are randomly shuffled.
Note that $H[s]$ and $H[k]$ are normalized by $\log{M}$, where $M$ is the size of the data.
For comparison to the unique data clustering of deep learning, the two entropies are also obtained
for data clustering by the k-means clustering algorithm, where k $\in \{2^{2}, \cdots, 2^{14} \}$ (filled orange triangles).
Theoretically maximal $H[k]$ for given $H[s]$ is also plotted (black dotted line).
(b) Degeneracy $m_k$ of the state frequency $k$ in the hidden layers of DBN epoch 200:
$H_3$ (empty black circles), $H_6$ (squares), and $H_9$ (triangles); and k=$2^{12}$ for k-means clustering (filled orange triangles).
(c) The power exponent in $m_k \propto k^{-\beta-1}$, for different layers at epoch 200 (filled black squares) and 0 (empty blue squares).
(d) Classification error for each hidden layer. 
Classification error is defined as the fraction of input samples that have the same hidden state but have different true labels from a majority true label for the hidden state. 
Ensemble averages of twelve realizations of the DBN were used for the plots and standard error estimation.
Errorbars are smaller than the symbols (c, d).
}
\end{figure*}

On the basis of the state and frequency entropies $H[s]$ and $H[k]$,
we examined the data grouping of the MNIST data~\cite{lecun1998mnist}.
The data contain $M=60,000$ samples of hand-written digits.
Each sample represents a $28 \times 28$ pixel image ($N=784$), where each pixel has a real value between 0 and 1.
Our DBN architecture has one visible ($V$) and ten hidden ($H_1, \cdots, H_{10}$) layers
that have a decreasing number of nodes (784-500-250-120-60-30-25-20-15-10-5) from $V$ to $H_{10}$.
After learning was completed, we obtained $\{\bm{h}^{\mu}_\ell\}_{\mu=1}^M$, $\ell \in \{1, \cdots, 10\}$ by
propagating the input data $\{\bm{v}^{\mu}\}_{\mu=1}^M$ forward to the hidden layers.
Again, we emphasize that DL is an agglomerative data grouping, especially in the case of narrowing DBN architectures (Fig.~\ref{Figure1}c).
We computed $H[s]$ and $H[k]$ for the ten hidden layers (Fig.~\ref{Figure2}a).
As the layer size shrinks, the state entropy $H[s]$ decreases monotonically due to the dimension reduction.
By contrast, the frequency entropy $H[k]$ increases up to the eighth hidden layer $H_8$ and then decreases.
Figure~\ref{Figure2}a shows that DL achieves representations with $H[k]$ that almost saturates the maximal theoretical value derived in Ref. \cite{marsili2015} up to layer $H_6$. The values of $H[k]$ are significantly higher than those obtained from running the widely used k-means clustering~\cite{bishop2006pattern}.

Further evidence that DL extracts representations with maximal $H[k]$ is obtained by looking at the frequency degeneracy $m_k$ of the hidden states for different layers. Figure~\ref{Figure2}b shows that the frequency degeneracy always follows a power law, $m_k \propto k^{-\beta -1}$, with a different exponent within each layer (Fig.~\ref{Figure2}c). These peculiar distributions are expected in the representation that maximizes $H[k]$ constrained by a fixed $H[s]=R$~\cite{marsili2013,marsili2015}.
This can be easily shown by means of the method of Lagrange multipliers, upon maximising 
\begin{equation}
\label{eq:freeenergy}
{\cal F} = H[k] + \beta (H[s] - R) + \lambda (\sum_k k m_k - M)
\end{equation}
with respect to $m_k$. Note that the second constraint comes from the normalization condition.
The maximization condition ($\delta {\cal F} / \delta m_k$=0) leads to $m_k \propto k^{-\beta-1}$.
The power exponent $\beta$ corresponds to (minus) the slope of the $H[s]$ vs $H[k]$ curve in Fig.~\ref{Figure2}a.
It would be misleading to interpret $\beta^{-1}$ as an effective temperature, because in our case
low temperature (high $\beta$) states correspond to high energy $\langle E\rangle$, whereas high 
temperature correspond to low energies. Indeed $\langle S\rangle$ increases with $\langle E\rangle$ 
when $\beta>0$, but the relation is concave rather than convex. Rather, $\beta$ tunes the trade-off 
between resolution and similarity: large $\beta$ emphasises resolution, whereas small $\beta$ accentuates 
similarities. 
Figure~\ref{Figure2}b also shows that k-means clustering does not produce a power laws frequency distribution, 
implying that the occurrence of power law behaviour $m_k$ is an emergent property of the DL representation and not an intrinsic property of the data. 

Figure~\ref{Figure2}a also compares the behaviour of the DBN after learning with its behaviour before learning, 
on the $H[s] - H[k]$ plane (see ``DBN epoch $0$'' data). More precisely, we generated the hidden states $\{\bm{h}^{\mu}_\ell\}_{\mu=1}^M$ before learning with a random $\theta$, and observed that most of the layers concentrate either on the low or on the high end of the resolution spectrum. 
The data grouping before learning was, of course, erroneous (Fig.~\ref{Figure2}d).
Nevertheless, interestingly, those layers that happen to be at an intermediate value of $H[s]$ also feature
high values of $H[k]$ and a power law distribution of frequencies (see Fig. S1). 
This suggests that the main effect of learning is to organize the representations in the subsequent layers in a hierarchical manner, covering as densely as possible the interval of relevant $H[s]$ values. The fact that DL extracts features at many resolution scales, however, 
{depends on the fact that these features are present in the data, in the first place}. Indeed, when comparing our results 
with those obtained from running the same DBN on re-shuffled data (see ``shuffled MNIST'' in Fig.~\ref{Figure2}a), we found again that the emergent representations were mostly at low or high resolution. At intermediate values of $H[s]$, we found 
representations which were far from the optimal ones. 

Let us now turn to discuss the generative capabilities of the network within the framework just outlined. A key aspect is that 
efficient generalisation entails an optimal trade-off between resolution and accuracy: Shallow layers (e.g. $H_1$, $H_2$, and $H_3$ in our example) infer inputs at too high resolution and fail to detect similarities between them. Hence they generate too noisy outputs. 
By contrast, the resolution of deep layers (e.g. $H_{10}$) is too low to generate the full spectrum of variability of the training set. 
The optimal generative power is expected to be achieved at intermediate layers and the theory outlined so far provides a quantitative criterium to identify it. Indeed, resolution and accuracy are both measured in bits. For optimal representations, a decrease of $\Delta E$ bits in resolution corresponds to an increase of $\Delta S\simeq \beta\Delta E$ bit in accuracy. 
Decreasing resolution provides relevant information on features that define the input's similarity when $\beta>1$. But for $\beta<1$, the information gain on similarity does not compensate the loss in resolution. Therefore {\em layers with $\beta\approx 1$ are those that achieve the optimal trade-off between resolution and accuracy}. There one bit of resolution is turned into one bit of information on features that define inputs' similarity. This implies that layers that achieve optimal generalisation ability {\em i)} are those for which $H[s]+H[k]$ is maximal and {\em ii)} that they exhibit a Zipf's law frequency distribution $m_k\sim k^{-2}$ (i.e. $\beta=1$). 

\begin{figure*}[t]
\begin{center}
\includegraphics[width=18.cm]{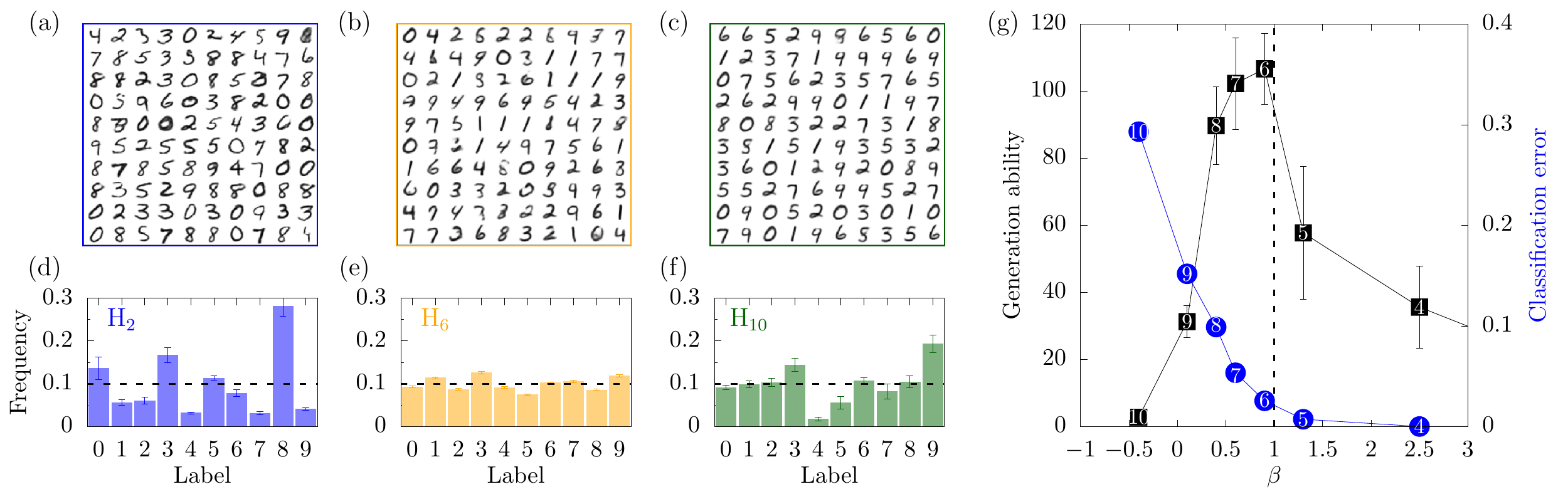}
\end{center}
\caption{\label{Figure3}
Optimal pattern generation of the critical layer.
Hand-written digit samples generated from (a) shallow ($H_2$), (b) critical ($H_6$), and (c) deep ($H_{10}$) hidden layers.
(d, e, f) Label frequencies of the generated samples. The dotted black lines represent the original label distribution of training samples.
(g) Generation ability (black squares) and classification error (blue circles) of the hidden layers (numbers in symbols).
The generation ability quantifies how closely the generated samples follow the statistics of the training samples (see the main text for details).
Here, the x-axis represents the power-law exponent $\beta$ in the degeneracy $m_k$ of the hidden state frequency $k$, $m_k \propto k^{-\beta-1}$, for each layer. Classification error is the same plot in Fig.~\ref{Figure2}d.
Twelve ensembles of generated samples were used to estimate the standard errors.
}
\end{figure*}

In order to test this claim, we examined the performance of the generative DBN.
After optimising $\theta$ so that the DBN learned the MNIST hand-written digits, we obtained equilibrium states for each hidden layer.
We repeated the backward and forward propagation between $H_{\ell}$ and $H_{\ell-1}$ 10,000 times for Gibbs sampling
to obtain the equilibrium states for the $\ell$th hidden layer, starting from 60,000 random initial states for $H_{\ell}$.
Then, we generated digit images in the visible layer $V$ by propagating the equilibrium states in $H_{\ell}$ all the way back to $V$.
The generated digits appeared different depending on the starting layer $H_{\ell}$ (Figs.~\ref{Figure3}a-c).
Shallow layer $H_2$ generated heterogeneous digit samples, including some odd-looking digits (Fig.~\ref{Figure3}a).
By contrast, deep layer $H_{10}$ generated stereotyped samples (Fig.~\ref{Figure3}c).
The generated digits mimic closely the original digits of the MNIST training samples. Yet, in order to evaluate the quality of generation, we decided to compare the statistics of generated and original samples, rather than their similarity, thus avoiding the choice of {\em ad-hoc} similarity measures. 
The DBN learned uniformly distributed digits (approximately 6,000 training samples for each digit, 0 to 9). 
We examined how closely the generated digits reproduce the original distribution of the training digit samples. In order to do this, we resorted to a classification machine for hand-written digits that works with an accuracy of 1.6\% error~\cite{hinton2006fast}, in order to assign labels to generated digits. 
Then, we quantified the {\em generation ability} as the inverse of the Kullback-Leibler divergence
\begin{equation}
\frac{1}{D(P||Q)} \equiv \bigg[\sum_{\text{label}=0}^9 P(\text{label}) \log \frac{P(\text{label})}{Q(\text{label})} \bigg]^{-1}
\end{equation}
between the label distribution of the training sample $P(\text{label})$ and that of 
the generated samples $Q(\text{label})$ (Figs.~\ref{Figure3}d-f).

Shallow layers $H_1$, $H_2$, and $H_3$ generated rich digit samples, but their label distribution $Q(\text{label})$ deviated substantially from $P(\text{label})$.
Shallow layers had a high resolution $H[s]$ but a low value of $H[k]$ (Fig.~\ref{Figure2}a).
Every training sample was mostly represented as distinct states and this precluded the shallow layers from extracting significant structures from the data.
On the other hand, deep layer $H_{10}$ generated stereotyped samples,
and their $Q(\text{label})$ also deviated from $P(\text{label})$.
Deep layers have low $H[s]$ and low $H[k]$, and fail to correctly distinguish different digits.
Indeed, deep layers have also a large classification error (Figs.~\ref{Figure3}g), because input samples with different labels should be represented by the same hidden states.
Finally, layer $H_6$, which achieved the highest value of $H[s]+H[k]$ and whose frequency distribution 
has a power law behaviour with $\beta\simeq 1$, was found to have the best generation ability and a low classification error (Fig.~\ref{Figure3}g). 
The highest $H[s]+H[k]$ may provide the largest flexibility for determining the number and size of distinct states,
which can contribute to effectively extract the nested data structures because it imposes least constraints for grouping similar samples with various sizes and separating odd samples as outliers.

We investigated different DBN architectures and datasets in order to assess the generality of our conclusions.
In the different (deep and narrowing) network architectures with different numbers of nodes/layers, the highest generation ability was always observed at the critical layer for which $H[s]+H[k]$ is maximal and whose frequency distribution is close to Zipf's law ($\beta=1$) (see Fig. S2).
Moreover, we confirmed that this conclusion was also true for using the images of lowercase letters in the OCR data~\cite{ocrdata}, as shown in the hand-written digit images (see Fig. S3).

\section{Discussion}
We characterise feature extraction in deep neural networks in terms of resolution and accuracy. The former captures information costs of the internal representations of training data, whereas the latter is expressed in terms of the degeneracy of energy (i.e. information cost) levels and it coincides with a recently proposed notion of relevance~\cite{marsili2015}. 
Within this picture, we found that DL achieves efficient data representations with maximal relevance at each level of resolution. Interestingly, maximal relevance has been shown to be an efficient criterium for extracting relevant variables in high dimensional data analysis also in other contexts~\cite{grigolon2016,cubero2018}. 
The resolution of the representation at each layer of the architecture is determined in an unsupervised manner, depending on the data. Indeed DL ``finds'' only features at high or low levels of resolution in structureless (i.e. random) datasets. 

A distinguishing feature of representations  with maximal relevance is that frequency distributions follow power laws. In particular, we argue that most efficient representations in generalisation power, are those that maximise the sum of relevance and resolution $H[s]+H[k]$, and have a frequency distribution that follows Zipf's law. Our numerical experiments fully corroborate this conclusion.
This suggests a further explanation for the ubiquitously observation of Zipf's law in nature~\cite{newman2005}, which is in line with arguments related to  information processing in language and communication evolutions~\cite{i2003least, hidalgo2014}, but which we believe is of a more fundamental nature. 

Our theory can be cast in the statistical physics framework suggested in~\cite{mora2011,Lee2015}. Yet, we believe that this analogy can be misleading in the present case. Defining energy as the negative of the logarithm of the frequency of a state alludes to a maximum entropy principle which may not be appropriate for learning machines. Indeed, we show that a principle of {\em minimum} entropy (i.e. maximal relevance) is more natural in this context. The relation between energy and entropy, indeed has the opposite convexity than in physics and, as a result, the parameter $\beta$ can hardly be interpreted as the inverse of a temperature. Also, the occurrence of Zipf's law and of power laws, while reminiscent of critical phenomena in physics~\cite{mora2011,tkavcik2015}, can arise in much more general settings (see e.g.~\cite{schwab2014}). 

Our findings extend to a range of network architectures and machine learning models beyond the deep belief networks discussed here. First, we confirmed that deep and gradually narrowing architectures were necessary to achieve high generation ability.
Deep but not narrowing and shallow networks showed poor generation ability, because they could not effectively extract the nested hierarchical structure in data (see Fig. S2). Interestingly, those networks still showed power-law-like frequency distributions of hidden states, but $H[k]$ stayed low compared to the deep and gradually narrowing networks.
We also examined sparse networks in terms of connectivities or activities.
The frequency distribution of their hidden states also followed power laws, 
although too strong sparsity lead to small $H[s]$ and $H[k]$ as well (Fig. S4). 
Second, we probed whether the power-law distributions were produced by discriminative models such as multi-layer perceptrons (MLP) and convolutional neural networks (CNN)~\cite{Lecun2015}.
Since their hidden states take real values unlike Boltzmann machines, we binarized the values of hidden variables.
The frequency distributions of binarized hidden states showed power laws in MLP and CNN (Fig. S5).
Thus, the emergence of power-law frequency distributions seems to be a general phenomenon for feed-forward neural networks as well as Boltzmann machines, suggesting that the emergence of maximally relevant representations is a general principle. 
Again, we found that the exponent $\beta$ of the power-law distribution provides useful information on the performance of discriminative models. For layer with $\beta < 1$ we found significant classification error after supervised learning. We confirmed that also variational autoencoder~\cite{kingma2013}, a  popular generative model, had also an optimal dimension for latent variables to achieve high generation ability (Fig. S6).

In summary, we suggest novel information theoretic measures to understand DL that we believe goes beyond the information bottleneck approaches~\cite{tishby2015deep, shwartz2017opening}. We hope this can contribute to derive a principled approach to design efficient network architectures for DL.

\bibliographystyle{unsrt}

\subsection{Acknowledgements}
We thank Saeed Saremi for the discussion during the initial stage of this study; and
Ji Hyun Bak and Changbong Hyeon for their helpful comments.
This research was supported by the Basic Science Research Program through the National Research Foundation of Korea (NRF), funded by the Ministry of Education (2016R1D1A1B03932264), and the Max Planck Society, Gyeongsangbuk-Do and Pohang City (J.J.), and partially supported by the ICTP through the OEA-AC-98 (J.S.).

\subsection{Author contributions}
J.S., M.M. and J.J. conceived the project; J.S. performed the simulations; all authors interpreted the results and contributed to the writing of the manuscript.

\subsection{Data availability}
All relevant data are available from the authors.

\newpage
\appendix
\section{Supplementary Information}

\section{Text S1. Simulation details}
The Boltzmann machine algorithm updates $\theta$ to maximize the log-likelihood of $\mathcal{L}(\theta)$ in Eq.~(5).
The updating formulation for the parameter is given by
\begin{equation}
\theta' = \theta + \frac{\alpha}{M} \frac{\partial \log \mathcal{L}(\theta)}{\partial \theta}
\end{equation}
with learning rate $\alpha$ given data size $M$.
The probability gradient for the coupling strength $W_{ij}$ is derived as
\begin{equation}
\label{bm}
\frac{\partial \log \mathcal{L}(\theta)}{\partial W_{ij}} = \sum_\mu \left[ \sum_{\bm{h}} v_i^\mu h_j P(\bm{v}^\mu, \bm{h}) - \sum_{\bm{v}} \sum_{\bm{h}}v_i h_j P(\bm{v}, \bm{h}) \right]. 
\end{equation}
The following probability gradients are obtained in a similar manner:
\begin{eqnarray}
\frac{\partial \log \mathcal{L}(\theta)}{\partial a_{i}} &=& \sum_\mu \left[ \sum_{\bm{h}} v_i^\mu P(\bm{v}^\mu, \bm{h}) - \sum_{\bm{v}} \sum_{\bm{h}}v_i P(\bm{v}, \bm{h}) \right], \nonumber \\
\\
\frac{\partial \log \mathcal{L}(\theta)}{\partial b_{i}} &=& \sum_\mu \left[ \sum_{\bm{h}} h_i P(\bm{v}^\mu, \bm{h}) - \sum_{\bm{v}} \sum_{\bm{h}} h_i P(\bm{v}, \bm{h}) \right]. \nonumber \\
\end{eqnarray}
To compute the gradients, one can use Gibbs sampling, called the contrastive divergence (CD) method~\cite{hinton2002},
instead of directly obtaining the joint probabilities, $P(\bm{v}, \bm{h})$.
A restricted Boltzmann machine(RBM) has a special network structure that the nodes in the same layer are not directly coupled. 
By the virtue of this network structure, the probability of visible/hidden node activity can be written as a product of the conditional probabilities of the individual nodes in a layer.

The conditional probability for forward propagation from $\bm{v}$ to the $j$th hidden node $h_j$ is
\begin{equation}
P(h_j | \bm{v}) = \frac{1}{1+\exp(\Delta E_j)},
\end{equation}
where $\Delta E_j \equiv E(\bm{v}, \bm{h})-E(\bm{v}, f_i(\bm{h}))$, and 
$f_j(\bm{h})$ is a flip operation for $h_j$.
Similarly, the conditional probability for backward propagation from $\bm{h}$ to the $i$th visible node $v_i$ is
\begin{equation}
P(v_i | \bm{h}) = \frac{1}{1+\exp(\Delta E_i)},
\end{equation}
where $\Delta E_i \equiv E(\bm{v}, \bm{h})-E(f_i(\bm{v}), \bm{h})$, and 
$f_i(\bm{v})$ is the flip operation for $v_i$.
We conduct Gibbs sampling with these conditional probabilities by propagating input data $\bm{v}^{\mu}(0)$ forward and backward $n$ times:
$\bm{v}^\mu(0) \mapsto \bm{h}^\mu(0) \mapsto \bm{v}^\mu(1) \mapsto \bm{h}^\mu(1) \mapsto \cdots \mapsto \bm{v}^\mu(n) \mapsto \bm{h}^\mu(n)$.
Then, the Gibbs sampling can approximate Eq.~(S2) as CD,
\begin{equation}
\label{eq:cd}
\frac{\partial \log \mathcal{L}(\theta)}{\partial W_{ij}} =  \sum_{\mu=1}^M v^\mu_i(0) h_j^\mu(0) - v_i^\mu(n) h_j^\mu(n).
\end{equation}
To find the global minimum more efficiently, we adopted the mini-batch method by using multiple batches of data
instead of considering all the data at once.
The randomly grouped batches introduce stochasticity to reduce the likelihood of becoming trapped in local minima.

The standard CD algorithm, however, can produce data-biased samples. 
To have more effective sampling, we adopted the persistent CD (PCD) method.
The initial state of the second term in Eq.~(\ref{eq:cd}) was random for the first batch, 
but final states ($\bm{v}^\mu(n), \bm{h}^\mu(n)$) for previous batches were used as the initial states for the following batches~\cite{tieleman2008training}.
We used 100 samples for a batch with $n=1$ PCD step for a batch, and computed the gradient in Eq.~(\ref{eq:cd}).
Then, we continued to compute the gradient with different batches. 
Each parameter update for a batch is called an epoch.
We used real mean values of activity from 0 to 1 for the visible node activities
and binary values (0 or 1) for the hidden node activities to improve the learning efficiency.
We updated $\theta$ for 200 epochs with a learning rate $\alpha = 0.1$.
The other details are described in \cite{keyvanrad2014brief}.

After the learning was completed, we obtained samples $\{ \bm{h}^\mu_{\ell} \}_{\mu=1}^M$ for the ${\ell}$th hidden layer 
by propagating input data $\{ \bm{v}^\mu \}_{\mu=1}^M$ forward to the deep layers. 
Since the Boltzmann machine is a stochastic machine, every realization has different $\{ \bm{h}^\mu_{\ell} \}_{\mu=1}^M$ given the same $\{ \bm{v}^\mu \}_{\mu=1}^M$. 
To obtain homogeneous (or reproducible) realization, we took dominant values of hidden activities by binarizing each component  $\langle{h_j} \rangle = P(h_j=1|\bm{v})$ of a hidden state $\bm{h}^\mu_{\ell}$ with a threshold of 0.5.
The binalization indeed corresponds to the zero temperature limit. 
In practice, after learning, most hidden activities are distributed near 0 or 1, so the binalization has little difference from the stochastic realization. However, before learning, the binalization sometimes lead to different statistics from the stochastic realization, because the hidden activities on non-optimized networks can be broadly distributed near 0.5.

In addition to the MNIST hand-written digit data, we applied our method to the OCR data~\cite{ocrdata}.
The data contain 52152 samples of lowercase letters. Each sample represents a $16 \times 8$ pixel image ($N=128$), 
where each pixel has a binary value (0 or 1). 
We used 44800 samples for training, 3720 samples to validate the hyperparameters, and 3632 samples for testing.
The original data contain the following frequencies of letters from a to z:
3333, 1148, 1849,  920, 4233,  721, 2441,  619, 3890,  164,  785, 2892, 1410, 4523, 3655, 1310, 74, 2584, 1075, 1696, 2373,  584,
 150,  392,  986,  and 990.
We used a discriminative machine to classify generated samples into lowercase letters.
We verified that the machine could successfully classify the original data, and found frequencies very close to the true values:
3294, 1140, 1847,  919, 4200,  801, 2380,  663, 4172,  189,  816, 2556, 1415, 4494, 3619, 1283,  140, 2570, 1090, 1672, 2258,  664,
 179,  413, 1033,  and 991.
Then, we assessed the generation performance, and confirmed that the critical layer ($\beta=1$) showed the highest generation ability
for the OCR data given a network architecture (128-320-160-120-100-95-90-85-75-70 for $V$, $H_1$, $\cdots$, $H_9$).
\bibliography{deep_clustering}

\begin{thebibliography}{10}

\bibitem{Byers2017}
Jeff Byers.
\newblock The physics of data.
\newblock {\em Nature Physics}, 13(8):718--719, 2017.

\bibitem{Lecun2015}
Yann LeCun, Yoshua Bengio, and Geoffrey Hinton.
\newblock Deep learning.
\newblock {\em Nature}, 521(7553):436--444, 2015.

\bibitem{Carrasquilla2017}
Juan Carrasquilla and Roger~G Melko.
\newblock Machine learning phases of matter.
\newblock {\em Nature Physics}, 13(5):431--434, 2017.

\bibitem{bengio2013representation}
Yoshua Bengio, Aaron Courville, and Pascal Vincent.
\newblock Representation learning: A review and new perspectives.
\newblock {\em IEEE transactions on pattern analysis and machine intelligence},
  35(8):1798--1828, 2013.

\bibitem{mehta2014exact}
Pankaj Mehta and David~J Schwab.
\newblock An exact mapping between the variational renormalization group and
  deep learning.
\newblock {\em arXiv preprint arXiv:1410.3831}, 2014.

\bibitem{tegmark2017}
Henry~W Lin, Max Tegmark, and David Rolnick.
\newblock Why does deep and cheap learning work so well?
\newblock {\em Journal of Statistical Physics}, 168(6):1223--1247, 2017.

\bibitem{bengio2009learning}
Yoshua Bengio et~al.
\newblock Learning deep architectures for ai.
\newblock {\em Foundations and trends{\textregistered} in Machine Learning},
  2(1):1--127, 2009.

\bibitem{chen2015deep}
Gang Chen.
\newblock Deep learning with nonparametric clustering.
\newblock {\em arXiv preprint arXiv:1501.03084}, 2015.

\bibitem{Note1}
The distribution $P(\protect \bm {h}|\protect \bm {v})$ of states in the hidden
  layer, for a fixed input, was found to be sharply peaked in all examples we
  studied, which makes it reasonable to approximate it with a singleton, as in
  hard clustering.

\bibitem{mora2011}
Thierry Mora and William Bialek.
\newblock Are biological systems poised at criticality?
\newblock {\em Journal of Statistical Physics}, 144(2):268--302, 2011.

\bibitem{marsili2013}
Matteo Marsili, Iacopo Mastromatteo, and Yasser Roudi.
\newblock On sampling and modeling complex systems.
\newblock {\em Journal of Statistical Mechanics: Theory and Experiment},
  2013(09):P09003, 2013.

\bibitem{marsili2015}
Ariel Haimovici and Matteo Marsili.
\newblock Criticality of mostly informative samples: a bayesian model selection
  approach.
\newblock {\em Journal of Statistical Mechanics: Theory and Experiment},
  2015(10):P10013, 2015.

\bibitem{lecun1998mnist}
Yann LeCun, Corinna Cortes, and Christopher~JC Burges.
\newblock The mnist database of handwritten digits.
\newblock http://yann.lecun.com/exdb/mnist/, 1998.

\bibitem{newman2005}
Mark~EJ Newman.
\newblock Power laws, pareto distributions and zipf's law.
\newblock {\em Contemporary physics}, 46(5):323--351, 2005.

\bibitem{hinton2006reducing}
Geoffrey~E Hinton and Ruslan~R Salakhutdinov.
\newblock Reducing the dimensionality of data with neural networks.
\newblock {\em Science}, 313(5786):504--507, 2006.

\bibitem{hinton2006fast}
Geoffrey~E Hinton, Simon Osindero, and Yee-Whye Teh.
\newblock A fast learning algorithm for deep belief nets.
\newblock {\em Neural computation}, 18(7):1527--1554, 2006.

\bibitem{huang2017accelerated}
Li~Huang and Lei Wang.
\newblock Accelerated monte carlo simulations with restricted boltzmann
  machines.
\newblock {\em Physical Review B}, 95(3):035105, 2017.

\bibitem{ackley1985}
David~H Ackley, Geoffrey~E Hinton, and Terrence~J Sejnowski.
\newblock A learning algorithm for boltzmann machines.
\newblock {\em Cognitive science}, 9(1):147--169, 1985.

\bibitem{bishop2006pattern}
Christopher~M Bishop.
\newblock {\em Pattern recognition and machine learning}.
\newblock springer, 2006.

\bibitem{ocrdata}
Rob Kassel.
\newblock Ocr dataset.
\newblock http://ai.stanford.edu/$\sim$btaskar/ocr/.

\bibitem{grigolon2016}
Silvia Grigolon, Silvio Franz, and Matteo Marsili.
\newblock Identifying relevant positions in proteins by critical variable
  selection.
\newblock {\em Molecular BioSystems}, 12(7):2147--2158, 2016.

\bibitem{cubero2018}
Ryan~John Cubero, Matteo Marsili, and Yasser Roudi.
\newblock Finding informative neurons in the brain using multi-scale relevance.
\newblock {\em arXiv preprint arXiv:1802.10354}, 2018.

\bibitem{i2003least}
Ramon~Ferrer i~Cancho and Ricard~V Sol{\'e}.
\newblock Least effort and the origins of scaling in human language.
\newblock {\em Proceedings of the National Academy of Sciences},
  100(3):788--791, 2003.

\bibitem{hidalgo2014}
Jorge Hidalgo, Jacopo Grilli, Samir Suweis, Miguel~A Mu{\~n}oz, Jayanth~R
  Banavar, and Amos Maritan.
\newblock Information-based fitness and the emergence of criticality in living
  systems.
\newblock {\em Proceedings of the National Academy of Sciences},
  111(28):10095--10100, 2014.

\bibitem{Lee2015}
Edward~D Lee, Chase~P Broedersz, and William Bialek.
\newblock Statistical mechanics of the us supreme court.
\newblock {\em Journal of Statistical Physics}, 160(2):275--301, 2015.

\bibitem{tkavcik2015}
Ga{\v{s}}per Tka{\v{c}}ik, Thierry Mora, Olivier Marre, Dario Amodei,
  Stephanie~E Palmer, Michael~J Berry, and William Bialek.
\newblock Thermodynamics and signatures of criticality in a network of neurons.
\newblock {\em Proceedings of the National Academy of Sciences},
  112(37):11508--11513, 2015.

\bibitem{schwab2014}
David~J Schwab, Ilya Nemenman, and Pankaj Mehta.
\newblock Zipf's law and criticality in multivariate data without fine-tuning.
\newblock {\em Physical review letters}, 113(6):068102, 2014.

\bibitem{kingma2013}
Diederik~P Kingma and Max Welling.
\newblock Auto-encoding variational bayes.
\newblock {\em arXiv preprint arXiv:1312.6114}, 2013.

\bibitem{tishby2015deep}
Naftali Tishby and Noga Zaslavsky.
\newblock Deep learning and the information bottleneck principle.
\newblock In {\em Information Theory Workshop (ITW), 2015 IEEE}, pages 1--5.
  IEEE, 2015.

\bibitem{shwartz2017opening}
Ravid Shwartz-Ziv and Naftali Tishby.
\newblock Opening the black box of deep neural networks via information.
\newblock {\em arXiv preprint arXiv:1703.00810}, 2017.

\bibitem{hinton2002}
Geoffrey~E Hinton.
\newblock Training products of experts by minimizing contrastive divergence.
\newblock {\em Neural computation}, 14(8):1771--1800, 2002.

\bibitem{tieleman2008training}
Tijmen Tieleman.
\newblock Training restricted boltzmann machines using approximations to the
  likelihood gradient.
\newblock In {\em Proceedings of the 25th international conference on Machine
  learning}, pages 1064--1071. ACM, 2008.

\bibitem{keyvanrad2014brief}
Mohammad~Ali Keyvanrad and Mohammad~Mehdi Homayounpour.
\newblock A brief survey on deep belief networks and introducing a new object
  oriented toolbox (deebnet).
\newblock {\em arXiv preprint arXiv:1408.3264}, 2014.

\bibitem{lee2008sparse}
Honglak Lee, Chaitanya Ekanadham, and Andrew~Y Ng.
\newblock Sparse deep belief net model for visual area v2.
\newblock In {\em Advances in neural information processing systems}, pages
  873--880, 2008.

\bibitem{lecun1998gradient}
Yann LeCun, L{\'e}on Bottou, Yoshua Bengio, and Patrick Haffner.
\newblock Gradient-based learning applied to document recognition.
\newblock {\em Proceedings of the IEEE}, 86(11):2278--2324, 1998.

\bibitem{tensorflow2015-whitepaper}
Mart\'{\i}n Abadi, Ashish Agarwal, Paul Barham, Eugene Brevdo, Zhifeng Chen,
  Craig Citro, Greg~S. Corrado, Andy Davis, Jeffrey Dean, Matthieu Devin,
  Sanjay Ghemawat, Ian Goodfellow, Andrew Harp, Geoffrey Irving, Michael Isard,
  Yangqing Jia, Rafal Jozefowicz, Lukasz Kaiser, Manjunath Kudlur, Josh
  Levenberg, Dandelion Man\'{e}, Rajat Monga, Sherry Moore, Derek Murray, Chris
  Olah, Mike Schuster, Jonathon Shlens, Benoit Steiner, Ilya Sutskever, Kunal
  Talwar, Paul Tucker, Vincent Vanhoucke, Vijay Vasudevan, Fernanda Vi\'{e}gas,
  Oriol Vinyals, Pete Warden, Martin Wattenberg, Martin Wicke, Yuan Yu, and
  Xiaoqiang Zheng.
\newblock {TensorFlow}: Large-scale machine learning on heterogeneous systems,
  2015.
\newblock Software available from tensorflow.org.

\bibitem{kingma2013auto}
Diederik~P Kingma and Max Welling.
\newblock Auto-encoding variational bayes.
\newblock {\em arXiv preprint arXiv:1312.6114}, 2013.

\end{thebibliography}

\setcounter{figure}{0}
\renewcommand{\thefigure}{S\arabic{figure}}

\begin{figure*}[h]
\begin{center}
\includegraphics[width=16cm]{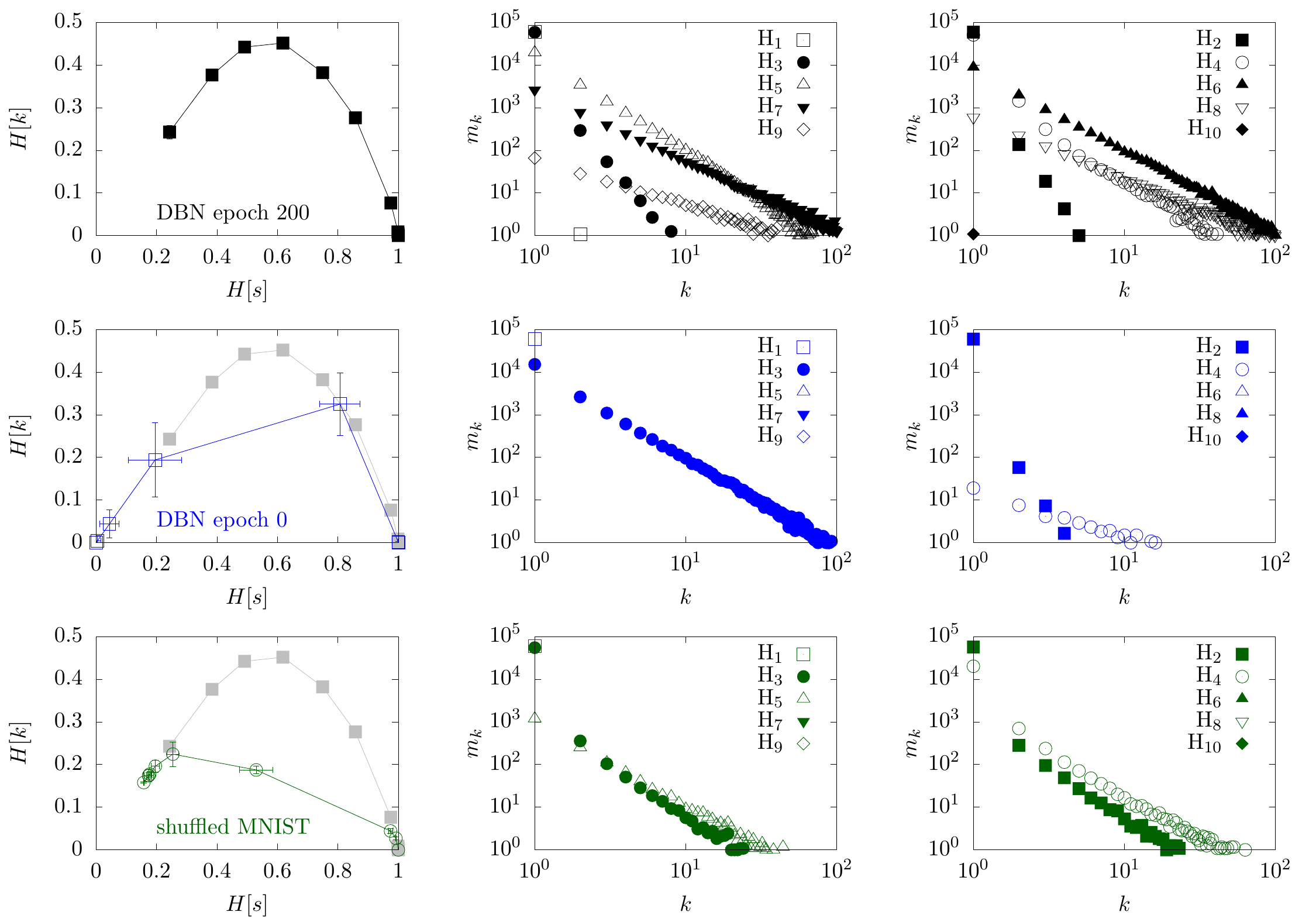}
\end{center}
\caption{ \label{Supp_Figure1}
Learning structureless data. State entropy $H[s]$ versus frequency entropy $H[k]$ for the MNIST data before learning (epoch 0), and after learning (epoch 200); and for the structureless data (shuffled MNIST). The shuffled MNIST data were generated by randomly switching black and white pixels in MNIST digit images.
For the learning, the standard deep network (500-250-120-60-30-25-20-15-10-5 nodes) was used.
The result of MNIST epoch 200 (gray filled square and line) is added for comparison.
Note that the uncertainty (standard deviation) of the two entropies for 12 ensembles decreased after learning for the MNIST data.
The corresponding frequency degeneracies $m_k$ for odd and even layers were separately plotted.
Some symbols out of the plot windows do not appear.}
\end{figure*}

\begin{figure*}[h]
\begin{center}
\includegraphics[height=13cm]{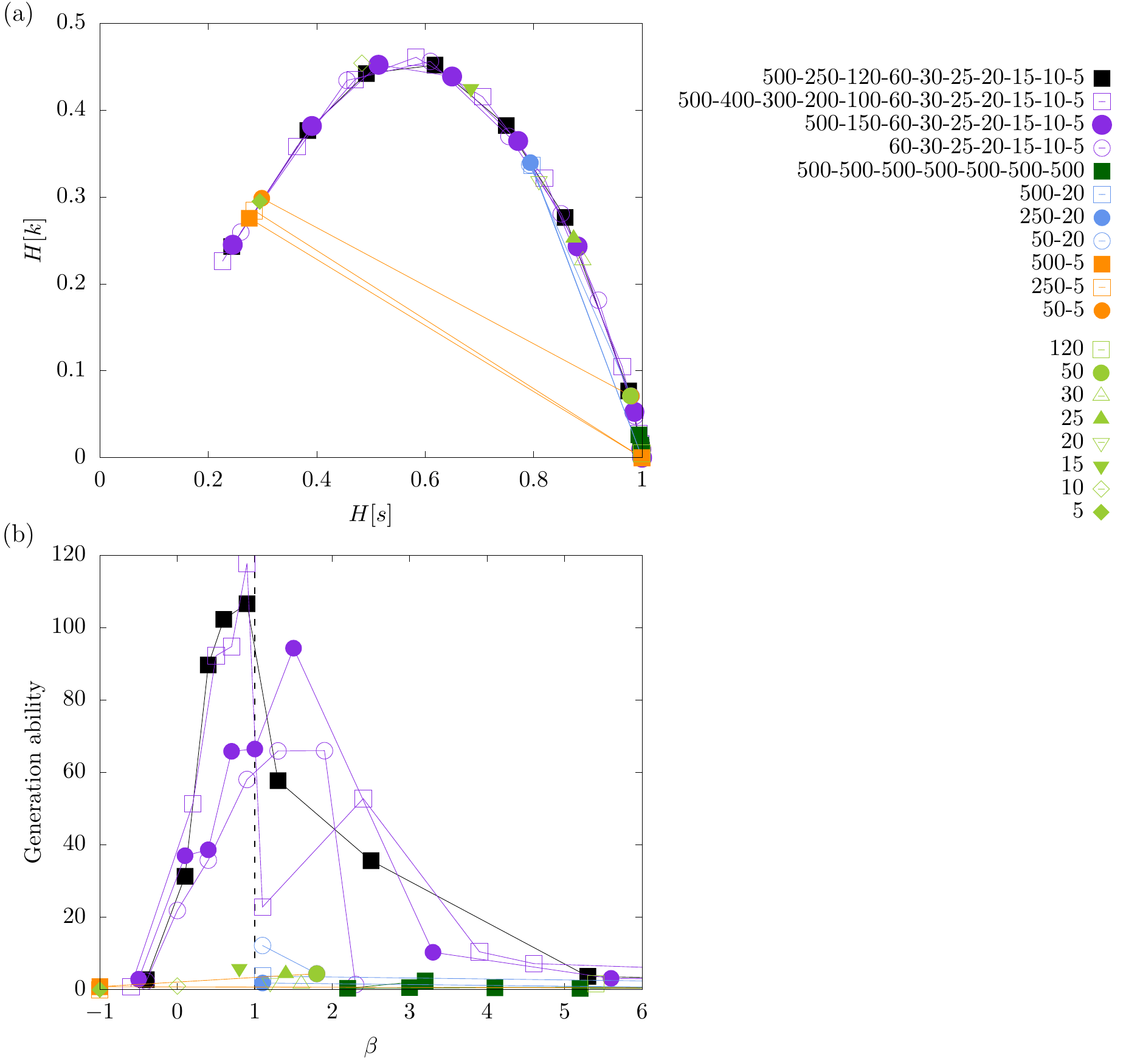}
\end{center}
\caption{ \label{Supp_Figure2}
Network structure and generation ability. (a) Two entropies $H[s]$ and $H[k]$ and (b) generation ability for various network structures:
deep and gradually narrowing structure; deep but not narrowing; shallow and narrowing; and shallow structures.
Note that the shallow networks correspond to the typical restricted Boltzmann machine (RBM) with a single hidden layer.
}
\end{figure*}

\pagebreak
\begin{figure*}[h]
\begin{center}
\includegraphics[width=16cm]{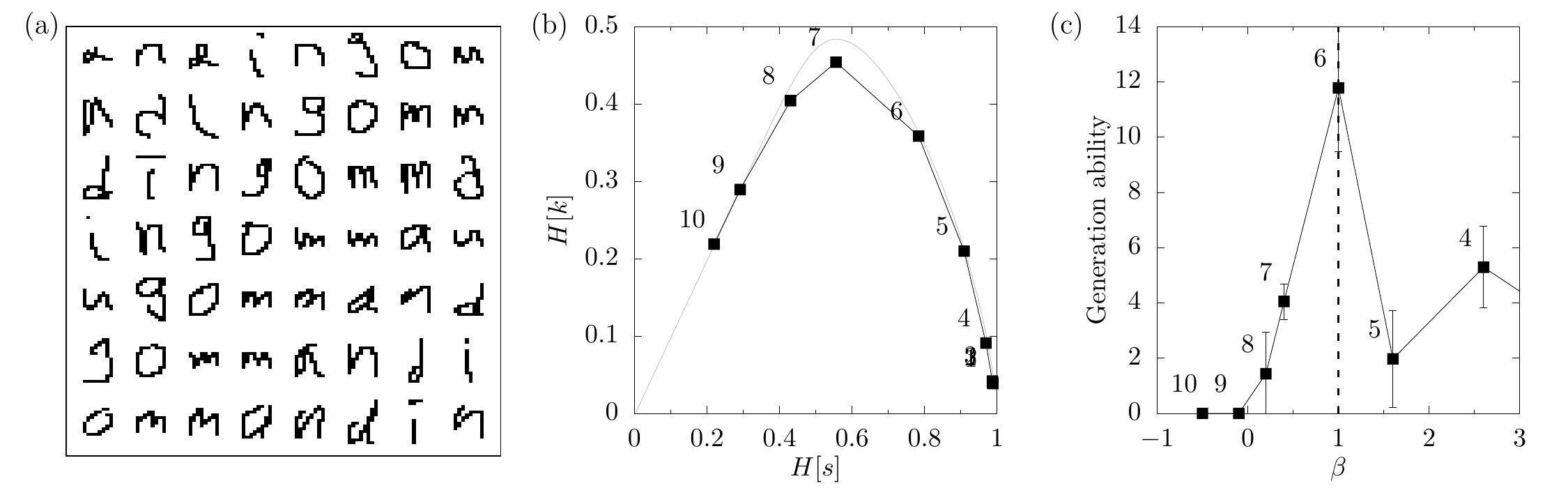}
\end{center}
\caption{ \label{Supp_Figure3}
OCR data. (a) 56 Lowercase letter images of the optical character recognition (OCR) data (16 $\times$ 8 pixels for each sample).
(b) Two entropies $H[s]$ and $H[k]$ represented on a deep neural network {(320-160-120-100-95-90-85-80-75-70}).
The gray line represents the theoretical maximal $H[k]$ for given $H[s]$.
(c) Generation ability of the ten hidden layers. Numbers beside symbols represent layer numbers.
The black dotted vertical line represents the critical layer ($\beta=1$).
Twelve ensembles of generated samples were used to estimate the standard errors.
}
\end{figure*}


\begin{figure*}[h]
\begin{center}
\includegraphics[width=16cm]{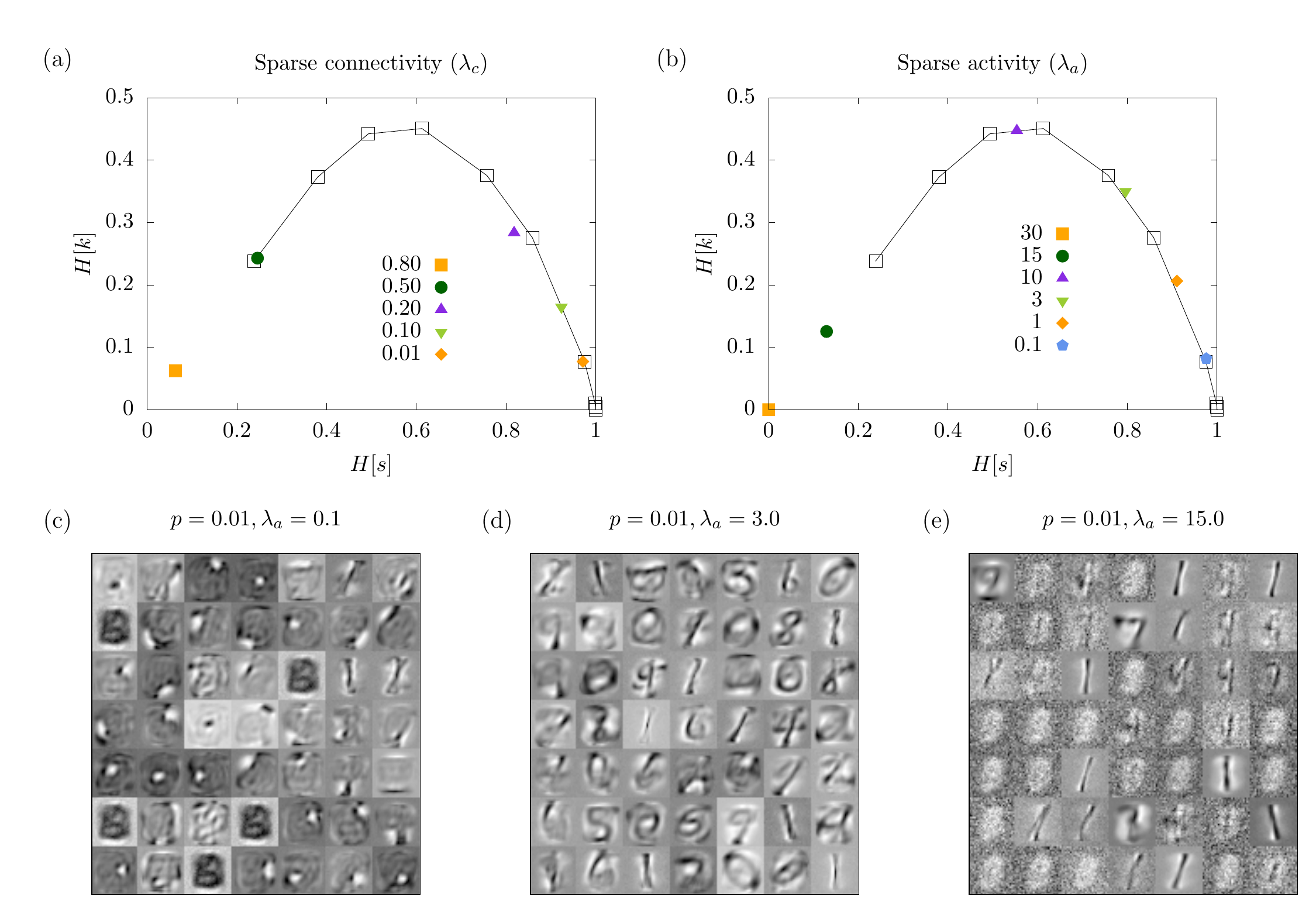}
\end{center}
\caption{ \label{Supp_Figure4}
Sparseness of network connectivity and activity. For a restricted Boltzmann machine (RBM) with 50 hidden nodes, we considered the sparseness of the network in terms of connectivities and activities.
To impose the sparseness, we used L2 regularization for hidden activities~\cite{lee2008sparse} as well as connection weights.
Then, the total cost included the negative log-likelihood and the two L2 penalties:
$C(\theta) = - \log L(\theta) + \lambda_c \sum_{i,j} W_{ij}^2 + \lambda_a \sum_{j} |p - M^{-1} \sum_{\mu=1}^{M} P\left(h_j^{\mu}=1|\bm{v}\right) |^2$.
Two entropies of $H[s]$ and $H[k]$ for (a) sparse connectivity ($\lambda_a= 0$) and (b) sparse activity ($\lambda_c= 0$).
(c, d, e) Features extracted from 49 hidden nodes (ignoring one hidden node for convenience of displays) under the sparse activity.
The features were obtained from the transpose of the weight matrices of each hidden node, $\bm{W}_j^{\top}$. 
We normalized each of them to have real values between 0 and 1, and then plotted them.
}
\end{figure*}

\begin{figure*}[h]
\begin{center}
\includegraphics[width=16.cm]{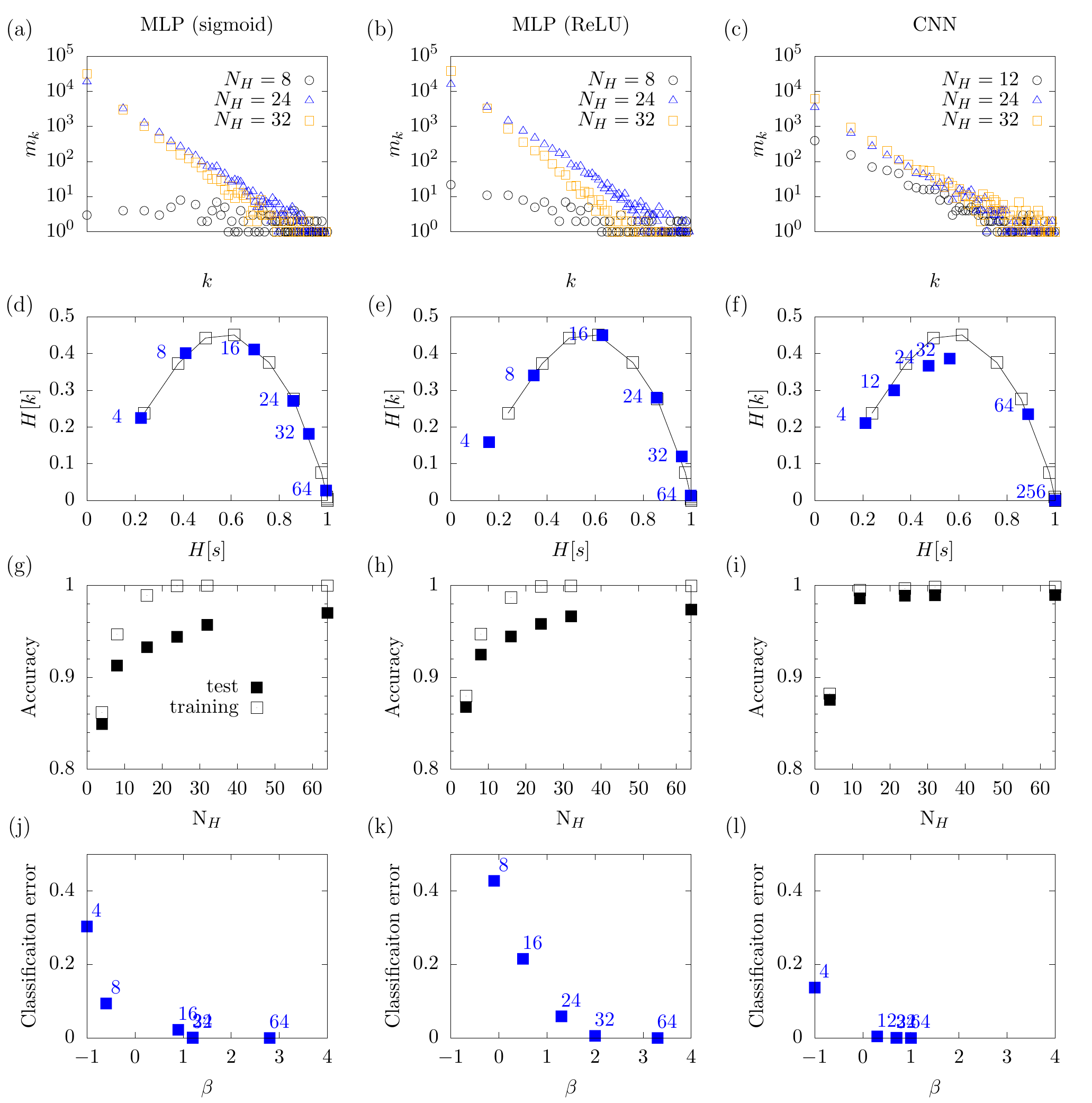}
\end{center}
\caption{ \label{Supp_Figure5}
Generalizability to discriminative models. Frequency degeneracies $m_k$ of binarized hidden states for (a) multi-layer perceptrons (MLP) using a sigmoidal activation function, (b) MLP using the rectified linear unit (ReLU), and (c) convolutional neural networks (CNN)~\cite{lecun1998gradient}.
The MLP has 3 layers where input, hidden, and output layers have 784, $N_H$, and 10 nodes, respectively.
The CNN has 7 layers with input, $5 \times 5$ convolution,  $2 \times 2$ max-pooling, $3 \times 3$ convolution,  $2 \times 2$ max-pooling, hidden, and output layers where the hidden layer with $N_H$ nodes is fully connected to the output layer with a sigmoidal activation function.
{Note that the plot is for the activities of last fully connected layer.}
(d, e, f) Two entropies of $H[s]$ and $H[k]$.
(g, h, i) Classification accuracy for training and test samples depending on the hidden-node number $N_H$.
(j, k, l) {\em Classification error} for power exponent $\beta$. 
Classification error is defined as the fraction of input samples that have the same hidden state but have different true labels from a majority true label for the hidden state.
For the classification of the MNIST hand-written digit images, we used the Python package of TensorFlow~\cite{tensorflow2015-whitepaper}
with the cross-entropy cost function, Adam optimizer, and Xavier initialization.
}
\end{figure*}

\begin{figure*}[h]
\centering 
\includegraphics[width=16cm]{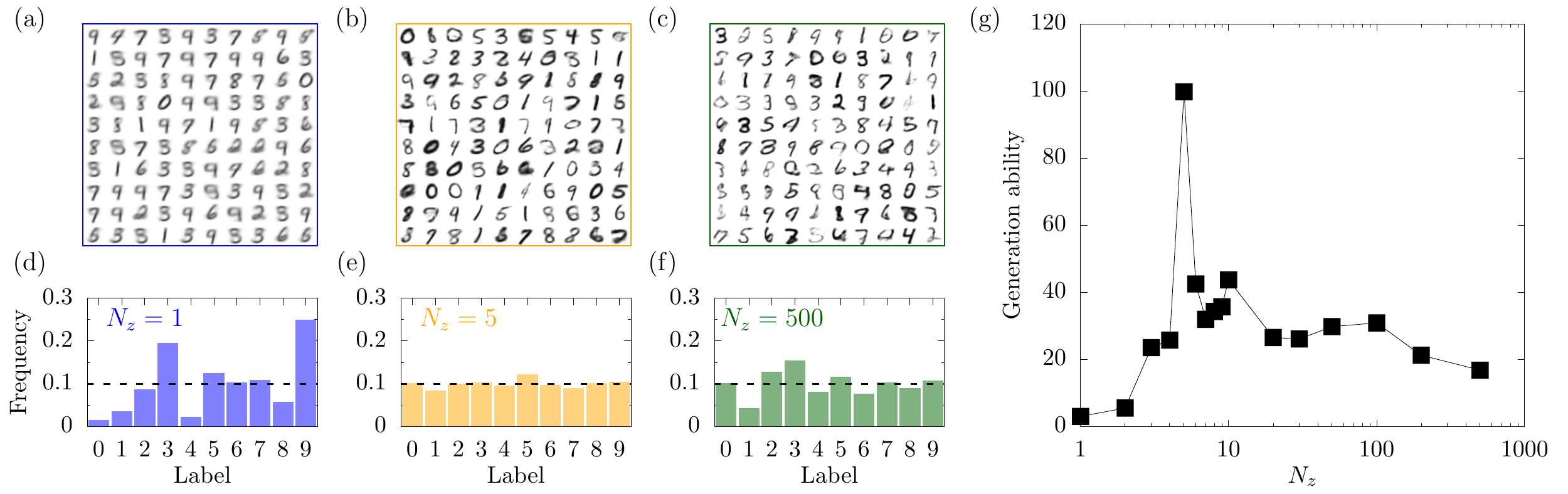}
\caption{\label{Supp_Figure}
Generation ability of variational autoencoder. Generated digit patterns by variational auto-encoders (VAE)~\cite{kingma2013auto} for the various dimensions of hidden (latent) variables: (a) $N_z=1$, (b) 5, and (c) 500.
The network structure of our VAE is 784-500(ReLU)-500(tanh)-$N_z$-500(tanh)-500(ReLU)-784(sigmoid).
(d, e, f) Label distributions of the generated patterns. Black dotted lines represent the original (uniform) distribution of MNIST digit samples.
(g) Generation ability (Eq. (9) in the main text) of the VAE depending on the dimension $N_z$ of the hidden (latent) variables.
}
\end{figure*}



\end{document}